\documentclass[a4paper, 10pt, conference]{ieeeconf}      %

\IEEEoverridecommandlockouts                              %
\makeatletter
\newcommand*\titleheader[1]{\gdef\@titleheader{#1}}
\AtBeginDocument{%
  \let\st@red@title\@title
  \def\@title{%
    \bgroup\normalfont\large\centering\@titleheader\par\egroup
    \vskip0.1em\st@red@title}
}
\makeatother

\usepackage{amsmath} %
\usepackage{amssymb}  %
\usepackage[textsize=footnotesize]{todonotes}
\usepackage{url} 
\usepackage{cite}
\usepackage{mathptmx} %
\usepackage[font=footnotesize, skip=3pt]{caption}
\usepackage[nodisplayskipstretch]{setspace}
\usepackage[textsize=tiny,final]{changes}

\usepackage{titlesec}
\titlespacing*{\section}{0pt}{*1}{*1}
\titlespacing{\subsection}{0pt}{*1}{*1}
\title{\LARGE \bf
	Teleoperation of Soft Modular Robots: Study on Real-time \\Stability and Gait Control 
}
\titleheader{\small{This paper has been accepted to 2023 IEEE-RAS International Conference on Soft Robotics (RoboSoft).}}

\author{Dulanjana~M.~Perera$^{1}$, Dimuthu~D. K.~Arachchige$^{2}$, Sanjaya~Mallikarachchi$^{2}$, Talal~Ghafoor$^{2}$,\\
	Iyad~Kanj$^{2}$, Yue~Chen$^{3}$, and Isuru~S.~Godage$^{4}$%
	\thanks{\!\!\!\!\!\!\!$^{1}$Department of Multidisciplinary Engineering, Texas A\&M University, College Station, TX 77843, USA.\newline $^{2}$College of Computing and Digital Media, School of Computing, DePaul University, Chicago, IL 60604, USA.\newline $^{3}$Department of Biomedical Engineering, Georgia Institute of Technology, Atlanta, GA 30332, USA.\newline $^{4}$Department of Engineering Technology \& Industrial Distribution and J. Mike Walker ’66 Department of Mechanical Engineering, Texas A\&M University, College Station, TX 77843, USA.
		\vspace{1mm}		
		\newline
		This work is supported in part by the National Science Foundation (NSF) Grants IIS–2008797, CMMI–2048142, and CMMI–2132994.
	}
}

\begin{document}
	
	\maketitle
	\thispagestyle{empty}
	\pagestyle{empty}

	\begin{abstract}
		Soft robotics holds tremendous potential for various applications, especially in unstructured environments such as search and rescue operations. However, the lack of autonomy and teleoperability, limited capabilities, absence of gait diversity and real-time control, and onboard sensors to sense the surroundings are some of the common issues with soft-limbed robots. To overcome these limitations, we propose a spatially symmetric, topologically-stable, soft-limbed tetrahedral robot that can perform multiple locomotion gaits. We introduce a kinematic model, derive locomotion trajectories for different gaits, and design a teleoperation mechanism to enable real-time human-robot collaboration. We use the kinematic model to map teleoperation inputs and ensure smooth transitions between gaits. Additionally, we leverage the passive compliance and natural stability of the robot for toppling and obstacle navigation. Through experimental tests, we demonstrate the robot's ability to tackle various locomotion challenges, adapt to different situations, and navigate obstructed environments via teleoperation.
		
	\end{abstract}
	
	\section{Introduction\label{sec:Introduction}}
	
	Robotics has had a significant impact on human civilization, with one key application being to replace humans in dangerous activities like search and rescue operations. Soft Robotics is a growing field that has made significant progress in design, modeling, and control in the past two decades~\cite{mazzolai2022roadmap,azizkhani2023dynamic}. Rigid robots have achieved advances in sensing, control, and navigation, but are limited by physical dimensions and cannot conform to the environment as well as soft robots. Soft robots' continuous deformation enables complex maneuvers in constrained spaces critical for inspection applications~\cite{bogue2019disaster} and better conformity to surroundings~\cite{schmitt2018soft}. Soft robots' inherent passive compliance mitigates the limitations of rigid robot control, paving the way for simpler robot designs and controllers. Rigid robot control tends to be more complex and requires multimodal sensory feedback to compensate for environmental interactions, while soft robots' compliance can simplify design and control~\cite{lee2017soft}.
	
	\begin{figure}[tb] 
		\centering
		\includegraphics[width=0.75\linewidth]{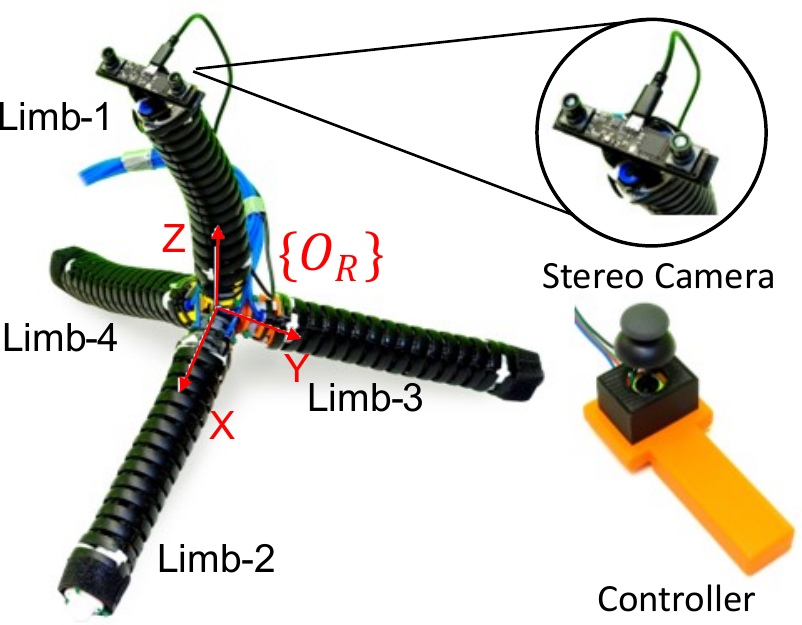}
		\caption{Tetrahedral robot with a stereo camera vision and proposed real teleoperation device.}
		\label{fig:1_IntroductionImage} 
	\end{figure}
	
	There are several soft robot designs inspired by biologies, such as the Eelworm robot that can crawl and swim on land~\cite{milana2020eelworm}, and soft robotic snakes that are capable of accessing tight spaces where legged robots cannot~\cite{arachchige2021soft}. However, legged robots have an advantage in navigating constrained or uneven terrain due to their ability to select contact points with their legs and utilize a variety of efficient gait patterns for different terrains, such as fast limb movements during the swing phase~\cite{wu2022fully, xia2021legged,godage2012locomotion}.
	
	Soft robots have shown remarkable terrainability and navigability in challenging environments, but they are not yet widely deployed in inspection, disaster response, and exploration applications. Several soft robotic prototypes have been proposed and evaluated for locomotion capabilities over different terrains~\cite{li2022scaling}, such as SoRX in \cite{liu2020sorx} and \cite{liu2021position}, and tortoise-inspired quadrupeds in \cite{mao2016design}, and \cite{huang2019soft}. However, these designs are limited by slow speeds and inefficient backward movements. The LEAP prototype, inspired by the cheetah, can achieve fast forward locomotion, turning, and climbing using electroadhesion \cite{tang2020leveraging}, while a similar gait was replicated in \cite{qin2019versatile}.
	
	Legged robot designs such as quadrupeds have limitations in stability during locomotion, especially in challenging terrains, and may topple. In contrast, the tetrahedral topology offers spatial symmetry that can be leveraged for stable locomotion and robust navigation. Wang et al.~\cite{wang2021design} proposed a soft-limbed tetrahedral robot that demonstrated fundamental locomotion gaits. However, the lack of proportional control in the limb actuators restricted the robot to a limited number of basic gaits, and its stability and robustness during locomotion were not thoroughly investigated.
	
	We present a novel tetrahedral soft robot with multiple gaits, achieved by using soft continuum modules that combine soft and rigid components to balance structural strength, compliance, and stiffness control~\cite{arachchige2022hybrid}. The modular design provides interchangeability, reliability, and robustness that are crucial for terrestrial robots subject to wear and potential damage. A complete kinematic model of the robot is derived, and a periodic limb motion trajectory is parameterized to achieve stable gaits for different locomotion modes. Gait control is proportional and dynamic, enabling robust navigation. The robot's topple-proof nature is investigated, and two approaches are proposed to recover from toppling from two aspects. One approach reorients the robot to the original pose whereas the other remaps the limb functionalities (a former limb becomes body, etc.) for continuing locomotion. Additionally, we demonstrate real-time human-robot collaboration through teleoperation in unstructured terrains.
	
	\begin{figure}[tb] 
		\centering
		\includegraphics[width=1.0\linewidth]{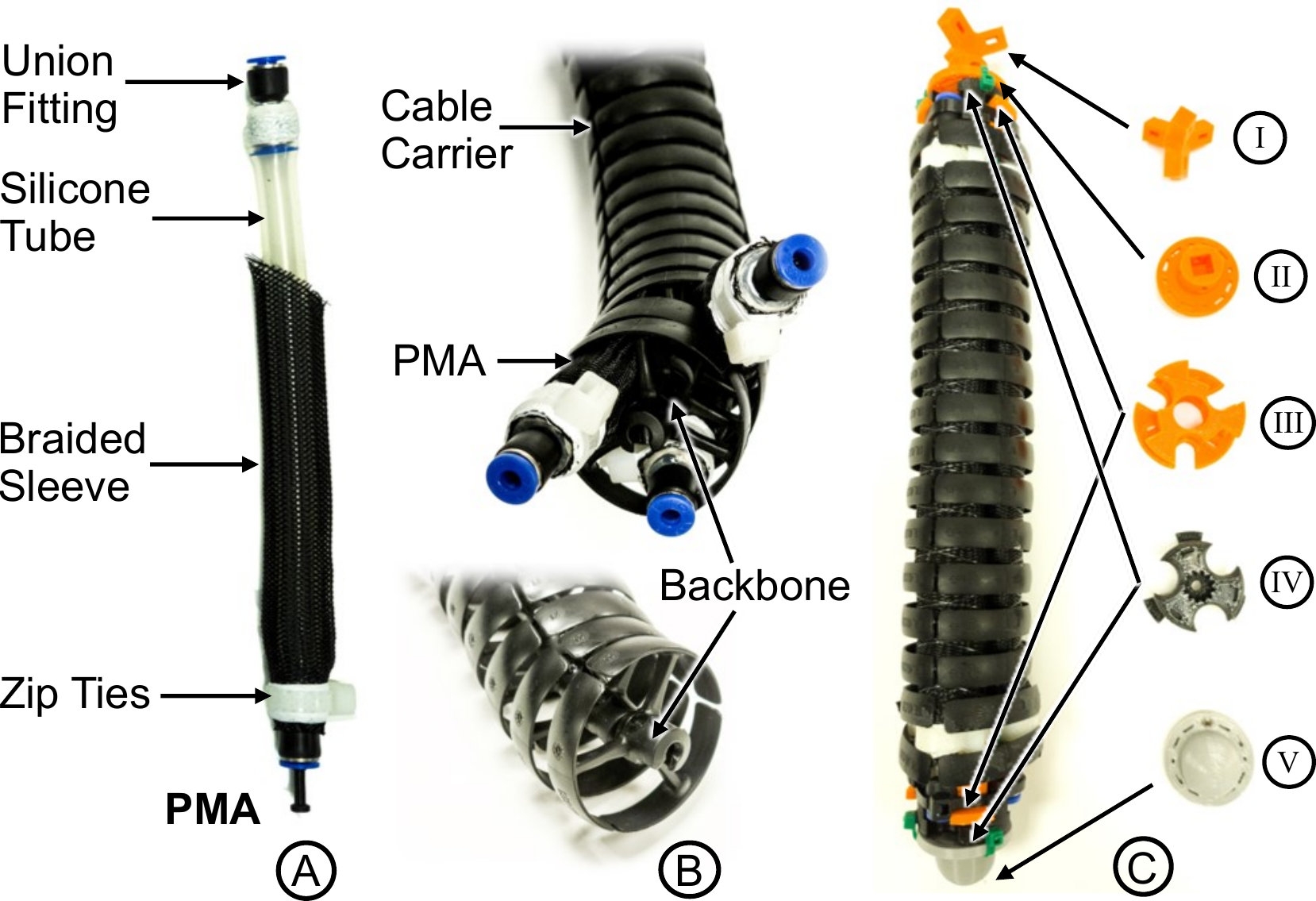}
		\caption{The design of a soft module. (A) Cross-section of a PMA. The top end of the braided sleeve is removed to illustrate the interior design. (B) Backbone structure and PMA placement inside the cable carrier. (C) External components. i) tetrahedral joint, ii) mounting bracket, iii) PMA fixture, iv) extension cap for external component connections, and v) compliant module end cap for uniform ground contact.}
		\label{fig:2_PartDesign} 
	\end{figure}
	
	\section{Prototype Description}\label{sec:PrototypeDescription}
	
	The soft robot proposed in this work is depicted in Fig.~\ref{fig:1_IntroductionImage}. The modular design approach simplifies fabrication, with each soft module consisting of a backbone formed by a commercially available cable carrier (Igus Triflex R-TRL40) and McKibben-type 
	extending pneumatic muscle actuators (PMAs) are fabricated using Silicone tubes and braided Nylon sleeves (Figs.~\ref{fig:2_PartDesign}A and \ref{fig:2_PartDesign}B). The backbone provides support for omnidirectional movements and sustains high forces and torques generated by the PMAs. The hybrid design methodology employed in the development of soft modules~\cite{amaya2021evaluation} results in better structural strength and stiffness--controlling capabilities required for locomotion. The tetrahedral topology of the robot provides natural stability in any orientation, making it more resilient to unexpected rolling, as discussed in Sec.~\ref{sec:Teleoperation}. The top soft module, or $Limb_1$, serves as the body limb and can be used to control the robot's center of gravity (CoG) during locomotion, as well as a sensing appendage for spatial data collection. 
	
	\begin{figure}[tb] 
		\centering
		\includegraphics[width=1.0\linewidth]{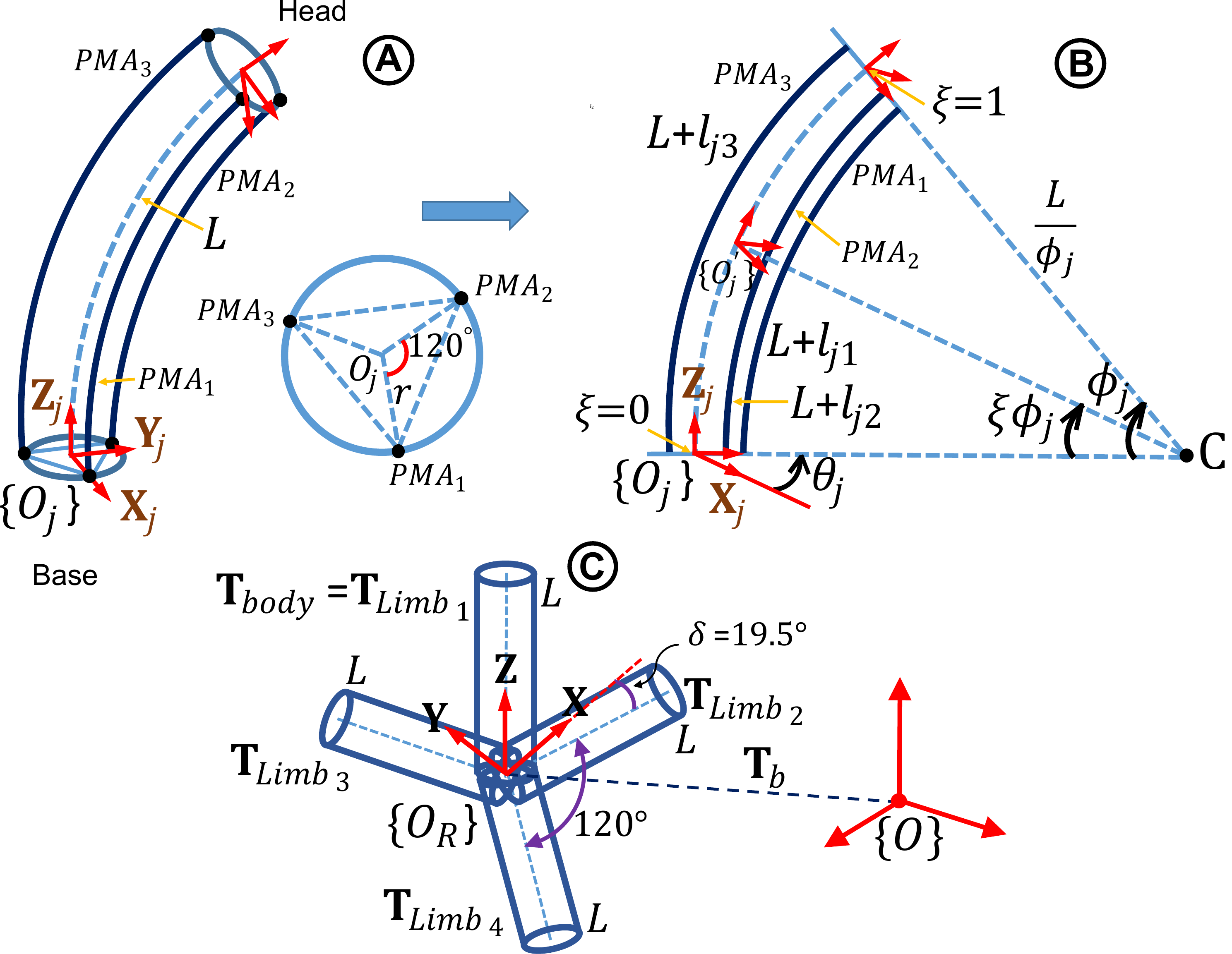}
		\caption{Schematic diagrams for kinematic modeling. (A) Isometric view of a single module and base plate description. (B) Kinematic details of the module. (C) Complete robot kinematic description}
		\label{fig:3_Schematicdiagram} 
	\end{figure} 
	
	The soft module, consisting of a backbone and pneumatic muscle actuators (PMAs), is designed to bend in a circular arc shape, as shown in Fig.~\ref{fig:2_PartDesign}C~\cite{deng2019near}. The rigid backbone, made of a commercially available cable carrier with a protective outer shell, constrains the length of the soft module, resulting in antagonistic operation of the PMAs that facilitates finer stiffness control over a wider range~\cite{arachchige2021novel}. The soft module has an effective length, diameter, and weight of $240~mm$, $40~mm$, and $0.15~kg$, respectively \cite{arachchige2022hybrid}. Four of these soft modules are connected via a 3D-printed tetrahedral joint to form the tetrahedral robot, which has a total of 12 actuated degrees of freedom (DoF) and weighs 
	$0.65~kg$ without the pneumatic pressure supply tubes. The 3D-printed parts used for mounting the PMAs in the grooves of the rigid chain are shown in Fig.~\ref{fig:2_PartDesign}C.
	
	\section{System Model}\label{sec:SystemModel}

	\begin{figure}[tb] 
		\centering
		\includegraphics[width=1.0\linewidth]{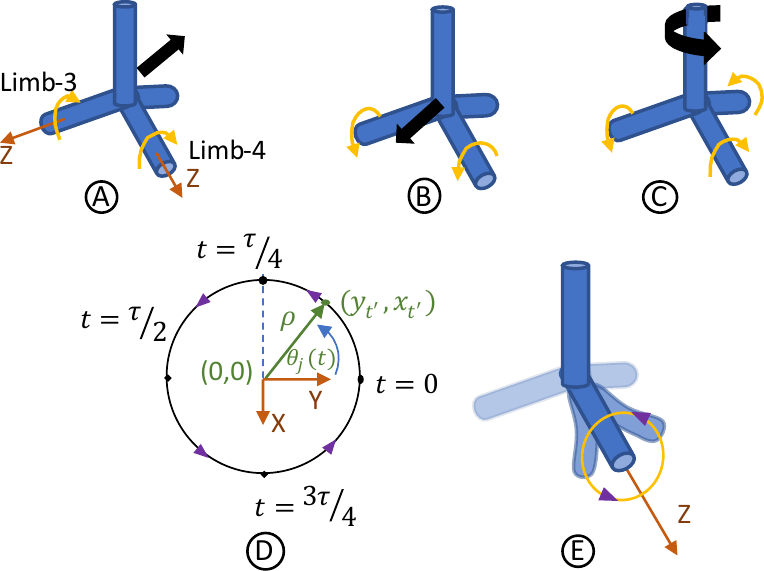}
		\caption{Locomotion Gait. (A) Limb motion for the forward movement. (B) Backward movement. (C) In-place turning. (D) XY plane projection of primary gait pattern for the forward movement (counter-clockwise direction).}
		\label{fig:4_gaitpattern} 
	\end{figure} 
	
	\subsection{Kinematics Modeling of Soft Modules}\label{subsub:KinematicsSoftLimbs}
	The workspace of a soft module is a surface symmetric about the Z-axis due to the inextensible backbone. The kinematic model of a soft module is derived by defining joint variables $l_{ji}\in \mathbb{R}$ as the length change of each PMA in the soft module, where $j$ is the module and $i\in \{1,2,3\}$ is the PMA index (Fig.~\ref{fig:3_Schematicdiagram}B). Configuration space variables are defined as the orientation angle $-\pi\leq\theta_{j}\leq\pi$ and the bending angle $0<\phi_{j}\leq\pi$ (see Fig.~\ref{fig:3_Schematicdiagram}). As shown in \ref{fig:3_Schematicdiagram}, due to the kinematic constraint imposed by the backbone, we have $l_{j1}+l_{j2}+l_{j3}=0$ at all times. Thus, one DoF out of the three actuated DoF is kinematically redundant due to the backbone constraint, which is utilized to achieve independent stiffness and shape control~\cite{deng2019near}. The relationship between the joint and configuration space variables is given by \eqref{eq:l2cp}, and the inverse relationship is given by \eqref{eq:cp2l}~\cite{arachchige2022hybrid,xiao2023kinematics}.
	\vspace{-1.5mm}
	\begin{align}
		l_{ji}& = -r\phi_{j}\cos\left(\textstyle\frac{2\pi}{3}\left(i-1\right)-\theta_{j}\right)  	\label{eq:l2cp}
	\end{align}
	where $r$ is the distance from the module center line to the PMA anchor points – module radius -- (Fig. \ref{fig:3_Schematicdiagram}A). 
	Note that, since air pressure is used to actuate the PMAs, $l_{ji}$ are converted into pressures ($bar$) using $P_{ji} = 80l_{ji} + 0.5$ and normalized to $0-3~bar$ which is the pressure range. 
	\vspace{-1.5mm}
	\begin{subequations}
		\begin{align}
			\phi_{j}&=\textstyle\frac{2}{r}\sqrt{\textstyle\frac{l^{2}_{j2}+l^{2}_{j3}+l_{j2}l_{j3}}{3}}\label{eq:cp_phi}\\
			\theta_{j}&=\arctan\left\{ \left(l_{j3}-l_{j2}\right),\sqrt{3}\left(l_{j2}+l_{j3}\right)\right\} \label{eq:cp_theta}
		\end{align}\label{eq:cp2l}
	\end{subequations}
	
	\begin{figure}[tb] 
		\centering
		\includegraphics[width=1\linewidth]{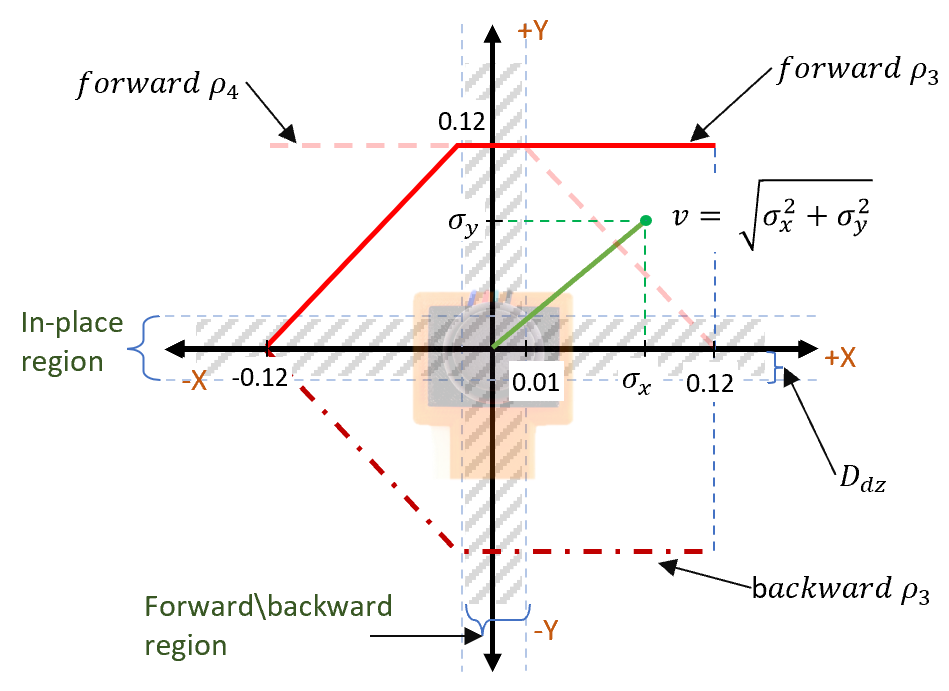}
		\caption{Gait-joystick map. The red lines indicate the variation of limb trajectory radius. The deadzone is indicated with a hatched region}
		\label{fig:gait-joystick} 
	\end{figure}
	\vspace{-3.5mm}
	We utilize the kinematic constraint $l_{j1}+l_{j2}+l_{j3}=0$ to eliminate the redundant variable $l_{j1}$ and simplify the relationships in \eqref{eq:cp2l}. Then, we perform a homogeneous coordinate transformation to map the configuration space--task space and obtain the spatial location of any point on the neutral axis (center line) of the module. The base coordinate frame ${O_j}$, point-frame ${O^\prime_j}$, and the respective transformation parameters are illustrated in Fig.~\ref{fig:3_Schematicdiagram}B, where the scalar $0 \leq \xi \leq 1$ represents any point along the soft module centerline. The complete homogeneous transformation matrix (HTM), $\mathbf{T}_{j}\in \mathbb{SE}(3)$, for the base-to-point of the $j$-th module is given by
	\vspace{-1.5mm}
	\begin{align}
		\mathbf{T}_{j}\left(\boldsymbol{q},\xi\right) & =\mathbf{R}_{Z}\left(\theta_{j}\right)\mathbf{P}_{X}\left(\textstyle\frac{L}{\phi_{j}}\right)\mathbf{R}_{Y}\left(\xi\phi_{j}\right)\mathbf{P}_{X}\left(\textstyle-\frac{L}{\phi_{j}}\right)\mathbf{R}_{Z}\left(-\theta_{j}\right)\nonumber \\
		& =\left[\begin{array}{cc}
			\mathbf{R}_j\left(\boldsymbol{q},\xi\right) & \mathbf{p}_j\left(\boldsymbol{q},\xi\right)\\
			0_{1\times3} & 1
		\end{array}\right] \label{eq:htm}
	\end{align}
	where $\mathbf{R}_{4\times4}\in \mathbb{SO}(3)$ and $\mathbf{P}_{4\times4}\in \mathbb{R}^{3}$ are standard homogeneous rotation matrices and translation matrices. The $q=[\theta_{j},\phi_{j}]^{T}$. The resulting position vector 
	$\mathbf{p}_{j}(q,\xi)$ or the forward kinematics, can be expressed as
	\vspace{-1.5mm}
	\begin{subequations}
		\begin{align}
			x_{j} & =L\phi_{j}^{-1}\cos\left(\theta_{j}\right)\left\{ 1-\cos\left(\xi\phi_{j}\right)\right\} \label{eq:x}\\
			y_{j} & =L\phi_{j}^{-1}\sin\left(\theta_{j}\right)\left\{ 1-\cos\left(\xi\phi_{j}\right)\right\} \label{eq:y}\\
			z_{j} & =L\phi_{j}^{-1}\sin\left(\xi\phi_{j}\right)\label{eq:z}	\end{align} \label{eq:xyz_tp}
	\end{subequations}
	\vspace{-3.5mm}
	
	For gait generation, inverse kinematic solutions are derived from  \eqref{eq:xyz_tp}. Even though $\theta_{j}$ can be analytically obtained, $\phi_{j}$ is solved numerically through an optimization process.
	\vspace{-1.5mm}
	\begin{subequations}
		\begin{align}
			\theta_{j} & =\arctan\left(y_{j},x_{j}\right)\label{eq:ik_theta}\\
			\textstyle \frac{1}{\phi_{j}}\left[1-\cos\left(\phi_{j}\right)\right] & =\textstyle \frac{1}{L}\sqrt{x_{j}^{2}+y_{j}^{2}}\label{eq:ik_phi}
		\end{align}
		\label{eq:xy_ik}
	\end{subequations}
	
	\subsection{Complete Kinematics of the Tetrahedral Robot}\label{subsub:CompleteKinematics}
	
	In the context of the proposed tetrahedral soft robot, a module is referred to as a limb. The floating-base coordinate frame, $\{O_R\}$, is fixed at the center of the tetrahedral robot aligning the local coordinate frame of $Limb_1$, as depicted in Fig. \ref{fig:3_Schematicdiagram}C. All subsequent coordinate transformations of limbs are obtained with respect to $\{O_R\}$. The complete HTM of each limb with respect to $\{O_R\}$ is given by
	\vspace{-1.5mm}
	\begin{subequations}
		\begin{align}
			\mathbf{T}_{Limb_1}\left({q},{\xi}\right) &=\mathbf{T}_{init}\left({q},{\xi}\right)\label{eq:Tlimb1}\\
			\mathbf{T}_{Limb_2}\left({q},{\xi}\right) &=\mathbf{R}_{Y}\left({\textstyle\frac{\pi}{2}+\delta}\right )\mathbf{T}_{Limb_1}\left({q},{\xi}\right)\label{eq:Tlimb2}\\
			\mathbf{T}_{Limb_3}\left({q},{\xi}\right) &=\mathbf{R}_{Y}\left({\textstyle\frac{\pi}{2}+\delta}\right )\mathbf{R}_{Z}\left({\textstyle\frac{2\pi}{3}}\right )\mathbf{T}_{Limb_1}\left({q},{\xi}\right)\label{eq:Tlimb3}\\
			\mathbf{T}_{Limb_4}\left({q},{\xi}\right) &=\mathbf{R}_{Y}\left({\textstyle\frac{\pi}{2}+\delta}\right )\mathbf{R}_{Z}\left({\textstyle\frac{4\pi}{3}}\right )\mathbf{T}_{Limb_1}\left({q},{\xi}\right)\label{eq:Tlimb4}
		\end{align} 
		\label{eq:T_robot}
	\end{subequations}
	\vspace{-3.5mm}
	
	From the tetrahedral geometry, we can obtain that $\delta =1.91- \frac{\pi}{2} \approx 0.34~rad$. In order to obtain the complete global kinematics, global transformation $\mathbf{T}_b(q_b)$ is defined as
	\vspace{-1.5mm}
	\begin{align}
		\mathbf{T}_b(q_b) &= \left[\begin{array}{cc}
			\mathbf{R}_b\left({q_{b}}\right) & \mathbf{p}_b\left({q_{b}}\right)\\
			0_{1\times3} & 1
		\end{array}\right]\label{eq:Tb}
	\end{align}
	where, $q_{b}=[\alpha, \beta, \gamma, x_b, y_b, z_b]^T$ are the global transformation parameters where $[\alpha, \beta, \gamma]$ and $[x_b, y_b, z_b]$ denote the Euler-angle offsets between coordinate frames $\{O\}$ and $\{O_{R}\}$ and the translation vector, respectively (see Fig. \ref{fig:3_Schematicdiagram}C).
	
	The complete kinematic model with reference to the global coordinate frame is given by
	\vspace{-1.5mm}
	\begin{align}
		\mathbf{T}_{Limb_j}\left({q}_{b},{q_j},{\xi}\right) &= \mathbf{T}_b\left ( q_{b} \right ) \mathbf{T}_{Limb_j} \left({q_j},{\xi}\right)\label{eq:Tcomplete}
	\end{align}

	\section{Derivation of Locomotion Gaits\label{sec:GaitGeneration}}
	
	\begin{figure}[tb] 
		\centering
		\includegraphics[width=1\linewidth]{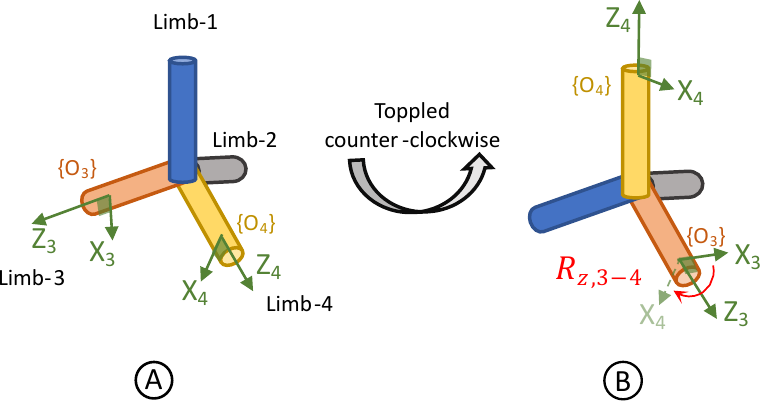}
		\caption{Toppling. (A) Starting orientation. (B) Orientation changed (topple). Limb-4 became Limb-1 and Limb-3 became Limb-4 and likewise, for others The light shade color represents the transformed axis.}
		\label{fig:toppleproof} 
	\end{figure}
	
	The pinniped locomotion gaits of a tetrahedral soft robot provide a more stable type of crawling motion due to having only three supporting limbs \cite{wang2021design}. All gaits, except turning in place, are derived from the fundamental crawling motion by synchronizing the limb motion with three parameters and mapped to teleoperation commands for real-time, continuous speed and direction control.
	
	\begin{figure}[tb] 
		\centering
		\includegraphics[width=1.0\linewidth]{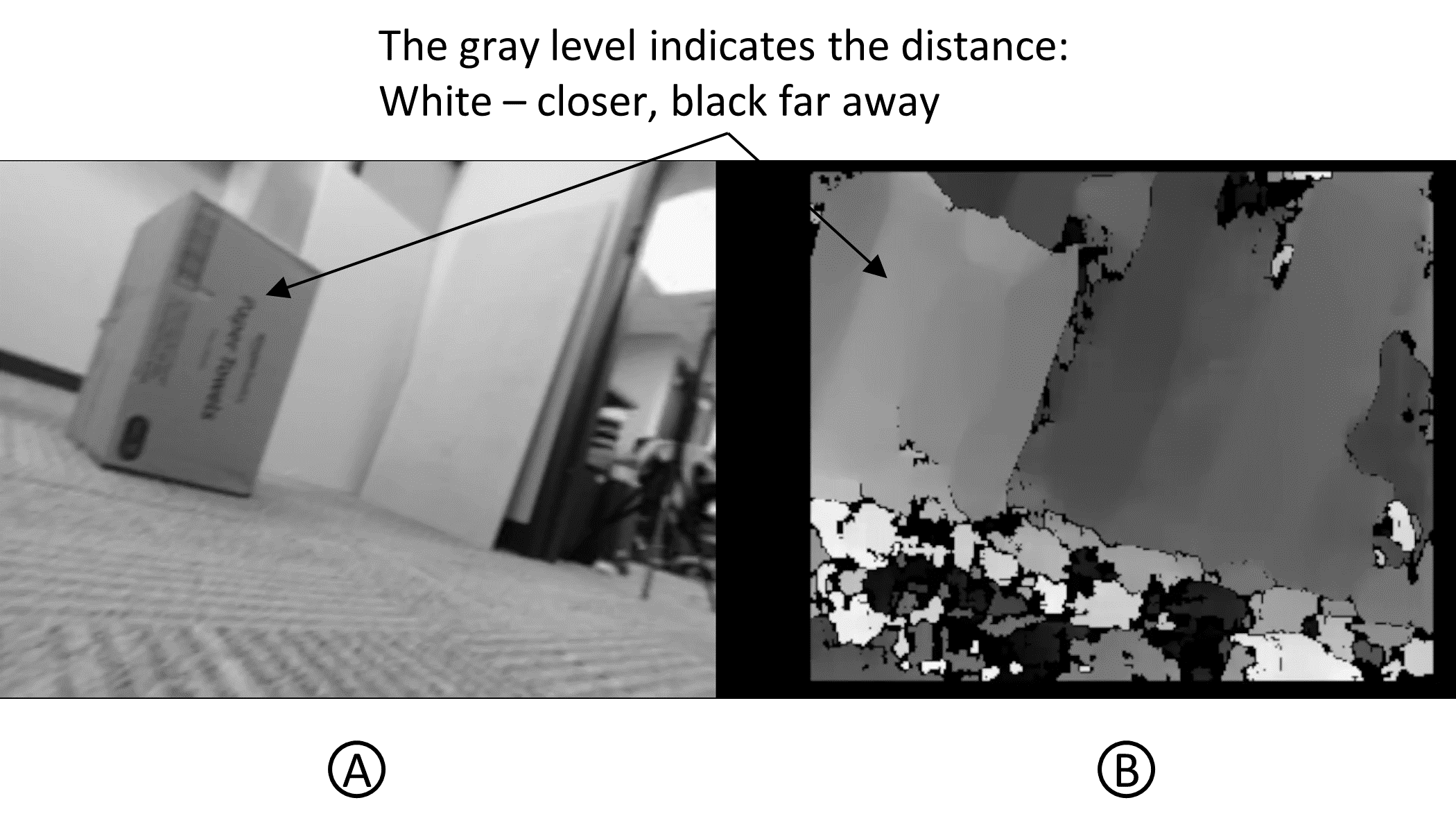}
		\caption{(A) Actual image from the camera at the robot. (B) Real-time depth map. Here gray level indicates the distance. The closer the object, the lighter the color.}
		\label{fig:depth} 
	\end{figure}
	
	\subsection{Fundamental Limb Motion for Locomotion  \label{subsub:PrimaryMotions}}
	
	As shown in Fig.~\ref{fig:3_Schematicdiagram}C, the symmetry of the tetrahedral topology allows us to select any two limbs as the primary thrusters for locomotion (Fig.~\ref{fig:4_gaitpattern}A). Without losing generality, we define a fundamental limb motion-circular limb trajectory (shown in Fig. \ref{fig:4_gaitpattern}) -- that can be parameterized by the trajectory radius $\rho_{j}$ and speed $\tau$. The limb's task space is a symmetric surface that can be projected onto the X-Y plane such that we can define the limb trajectory in $(x_j, y_j)$ coordinates for the $j$-th limb. Joint variable trajectories are computed using \eqref{eq:xy_ik} and \eqref{eq:l2cp}. A phase-offset parameter $\beta_j$ synchronizes all limb movements. Note that, depending on the various phase offsets employed in limbs, the robot CoG may be shifted to reduce the drag resulting from the limb-ground contacts. Varying the radius $\rho \in \left[0,0.12\right]$ controls the speed of linear movement and turning. The gait period $\tau$ determines the trajectory speed per cycle and is fixed at $\tau = 100$ for all experiments.
	
	From geometry, we can compute the circular workspace  trajectory for a given $\rho_{j}$ as 
	\vspace{-1.5mm}
	\begin{subequations}
		\begin{align}
			x_j &=-\rho_{j} \cos 
			\left(
			\textstyle\frac{2\pi t}{\tau} +\beta_{j}
			\right)
			\\ 
			y_j &=\rho_{j} \sin
			\left(
			\textstyle\frac{2\pi t}{\tau} +\beta_{j}
			\right)
		\end{align}
		\label{eq:gaitXY}
	\end{subequations}
	$\!\!$where $\theta_{j}(t) = \frac{2\pi t}{\tau} \in [0,2\pi]$ and $t\in[0,~\tau]$ is the corresponding period parameter. The $-\rho$ indicates the rotation direction (counter-clockwise) as depicted in Fig. \ref{fig:4_gaitpattern}. The parameter $\beta$ defines the phase shift of the trajectory. It opens opportunities to alter the gait for different applications. However, in this study, a phase shift is only used to obtain mirrored movements for $Limb_3$ and $Limb_4$.
	
	\subsection{Forward and Backward Crawling With Turning \label{subsec:ForwardBackward}}
	
	The $+X$-axis is considered as the forward-moving direction (as in Fig.~\ref{fig:4_gaitpattern}A). To move forward, $Limb_3$ rotates anti-clockwise about $+Z$ of $\{O_3\}$ and clockwise about $+Z$ of $\{O_4\}$. Thus, the gait trajectory for $Limb_3$ is given by \eqref{eq:gaitXY} with $\beta_{3}=0$, and for $Limb_4$, the same equation with $\beta_{4}=\pi$ is used. For backward movement, $-\rho$ in \eqref{eq:gaitXY} is changed to $+\rho$, and $\beta_{3}=\pi$ and $\beta_{4}=0$ are set.
	
	\begin{figure}[t] 
		\centering
		\includegraphics[width=1\linewidth]{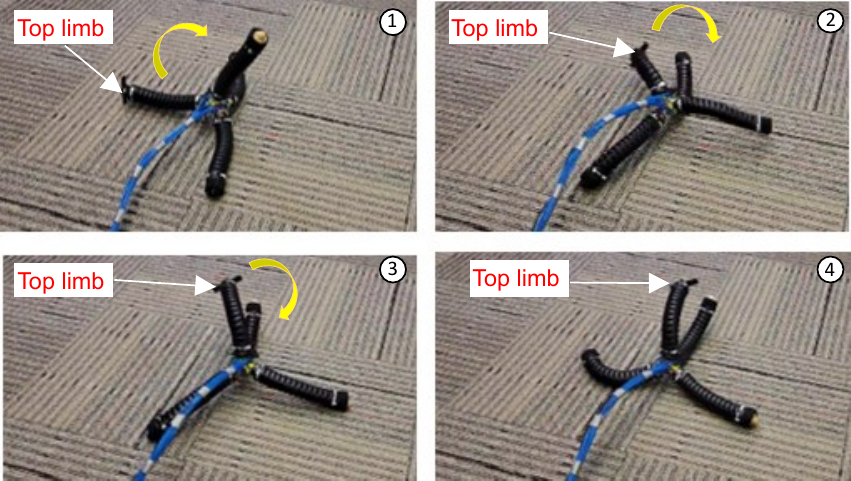}
		\caption{Self-orientation correction. The robot is toppled to the left side and a right turn is executed to correct the orientation.}
		\label{fig:selfcorrection} 
	\end{figure}
	
	\begin{figure*}[t] 
		\centering
		\includegraphics[width=1\textwidth]{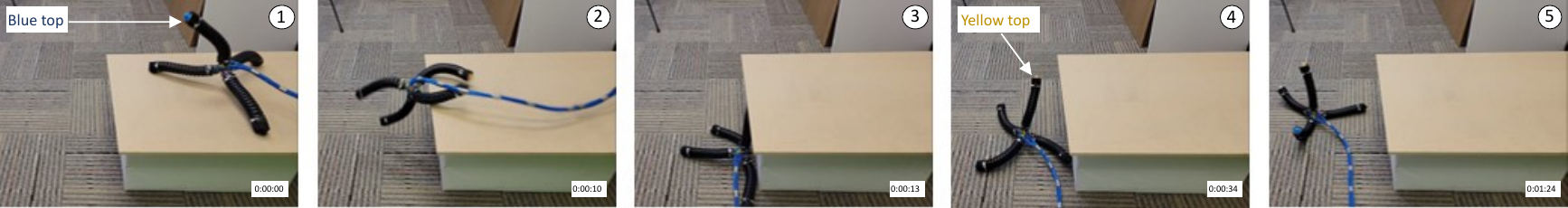}
		\caption{Topple proof demonstration. (1) Robot moves with a blue marker at the tip (i.e., end). (2) Falling. (3) Remap the limb configuration according to the Yellow limb. (4) Push away from the obstacle. (5) Moves forward as normal.}
		\label{fig:toppleproofexp} 
	\end{figure*}
	
	The movement \textit{turn-left} and \textit{turn-right} is achieved by altering the gait radius $\rho_j$ of the limbs. To turn right, for instance, the radius of $Limb_3$, $\rho_3$, is increased or maintained at a constant value (depending on the user's desired speed), while the radius of $Limb_4$, $\rho_4$, is reduced. This disparity in speed results in the robot turning right while simultaneously advancing forward, and similar principles apply to other turning movements. Thus, dynamic gait generation enhances the operator's steering capabilities during navigation. The connection between the operator's input on the console and gait trajectory generation is detailed in Sec.~\ref{sec:Teleoperation}.
	
	\begin{figure}[tb] 
		\centering
		\includegraphics[width=1\linewidth]{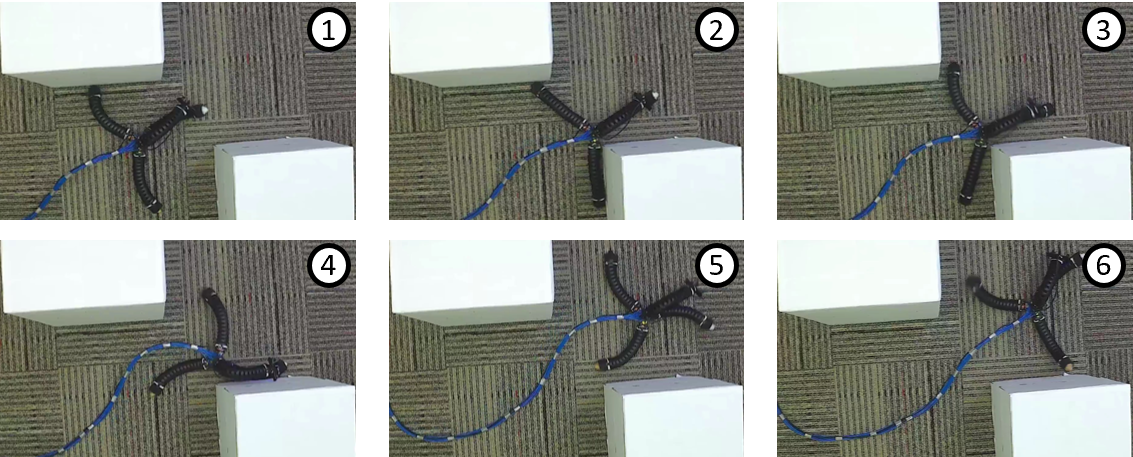}
		\caption{Limbs deform and allow the robot to squeeze through a narrow gap.}
		\label{fig:squeezing} 
	\end{figure}
	
	\subsection{In-place Turning \label{subsec:TurningIn-place}}
	In-place turning refers to the rotation around the $Z$-axis of the robot coordinate frame $\{O_R\}$ (as shown in Fig.~\ref{fig:4_gaitpattern}C). This gait is versatile for inspection tasks since the operator can turn the robot in both clockwise/counterclockwise without any lateral motion. To achieve this gait, the fundamental circular gait is executed on $Limb_2$, $Limb_3$, and $Limb_4$ using a calculated $\rho_j$ value obtained from the joystick. The left in-place turning is described by $x_j =\rho_{j} \cos(\frac{2\pi t}{\tau}),~y_j =\rho_{j} \sin(\frac{2\pi t}{\tau} +\pi)$, where $j={2,3,4}$, and similar principles apply for right-rotation, with $-\rho$ used to alter the direction of $x_j$.
	
	\subsection{Stability Maintain \label{subsec:stability}}
	Despite the inherent stability of tetrahedral topology and crawling movements in theory, uneven ground conditions may cause the robot to become unstable due to excessive shifting of its CoG. At rest, the CoG is at the center (origin of $\{O_{R}\}$). However, during high-speed crawling ($\rho>0.12$), the CoG may shift beyond the support region, where the reaction forces from the ground on $Limb_3$ and $Limb_4$ can cause a net torque and tipping during transverse movements. To mitigate this issue, we shift the robot CoG forward by bending the top limb ($Limb_1$) and increasing static stability. Note that it is possible to dynamically adjust the CoG and enhance stability for gaits other than crawling.
	
	\section{Teleoperation\label{sec:Teleoperation}}
	
	The main aim of this study is to investigate the real-time human-robot collaboration between an operator and the proposed soft tetrahedral robots for inspection operations in challenging spaces. 
	
	\subsection{Teleoperation Console Design and Development\label{subsec:remotecontroller}}
	
	The teleoperation console is intended to assist the operator in executing all crawling locomotion modes, such as moving forward/backward, turning, and in-place turning. It enables smooth transitions between gaits and provides proportional control over them. Furthermore, the system enables the user to choose the robot's orientation to execute a particular mechanism during toppling. The teleoperation system also leverages visual feedback from the robot's camera, which is processed to obtain depth information that helps the operator make informed decisions. 
	
	We employ a dual-axis joystick 
	to sense user inputs to the teleoperation console -- a 3D printed handheld enclosure (see Fig.~\ref{fig:1_IntroductionImage}). This joystick device is composed of two trimpots, each with a resistance of $10~k\Omega$, for the $x$ and $y$ axes. It has a lever range of $\pm 30^\circ$ a toggle press switch. The joystick provides a linear response within $\pm 300~\Omega$, and is not affected by deadzones. We interface the voltage responses of the two trimpots using an NI PCI-6221  DAQ card and a MATLAB Realtime Simulink model. The joystick output signals are in the range of $[1, 5]~V$ and are mapped to the fundamental limb trajectory radius, $\rho \in [0,~0.12]$. To smoothen user inputs, we introduce a deadzone around the origin for both axes, which is set at $\pm 0.01$ (see Fig.~\ref{fig:gait-joystick}). This enables the operator to execute stable, in-place turning and transverse motions without any interruptions.
	
	\subsection{Real-time Gait Mapping to User Inputs\label{subsec:Joystick}}
	
	The turning and transverse movements are dynamically combined to obtain high-fidelity composite gaits which facilitate turn while moving (Fig.~\ref{fig:gait-joystick}). When the joystick position for both limbs is within the defined deadzone, $D_{dz}$, along the $Y$-axis with the same value of $\rho$, the robot moves forward and backward without turning. To enable a seamless and continuous transition between axial movement and turning while moving, we use \eqref{eq:leftR} and \eqref{eq:rightR}. However, in-place turning is executed when the joystick position falls within $D_{dz}$ along the \textit{x-axis}. The value of $\rho$ is determined by \eqref{eq:speed}, which specifies the radial distance of the joystick position.
	\vspace{-1.5mm}
	\begin{subequations}
		\begin{align}
			v &= \sqrt{\sigma_{x}^{2} + \sigma_{y}^{2}}\label{eq:speed}\\
			\rho_{3} &= f\left ( \sigma_{x},v \right )\left ( \sigma_{x}+v \right ) f\left ( -\sigma_{x},D_{dz} \right ) \nonumber\\
			& \qquad +\left ( v-D_{dz} \right ) f\left ( \sigma_{x},D_{dz} \right )\label{eq:leftR}\\
			\rho_{4} &= f\left ( -\sigma_{x},v \right )\left ( -\sigma_{x}+v \right )f\left ( \sigma_{x},-D_{dz} \right )\nonumber\\
			&\qquad+\left ( v-D_{dz} \right )f\left ( -\sigma_{x},D_{dz} \right )\label{eq:rightR}
		\end{align}
	\end{subequations}
	where
	\begin{equation}
		f\left ( a,b \right) = \textstyle \frac{1}{2}\left[\tanh{\left [ \left(a + b\right)10^{6} \right]} + 1\right] \nonumber
	\end{equation}
	with $\sigma_{x},\sigma_{y}\in [-0.12,~0.12]$ are the joystick \textit{x-axis} and \textit{y-axis} values, respectively. The detailed gait-joystick map is depicted in Fig. \ref{fig:gait-joystick}
	
	\subsection{Realtime Stereo Computer Vision System \label{subsec:Cameramounting}}
	The symmetry of the tetrahedral topology enables uninterrupted and reorientation-free completion of field tasks. In theory, it is possible to install cameras at the end of each soft limb such that, even if the robot gets disoriented the cameras at other limbs can be utilized. To test the feasibility of real-time navigation via teleoperation using visual feedback, we use a single Hotpet Synchronized Dual Lens stereo camera in this study (as shown in Fig.~\ref{fig:1_IntroductionImage}). The two cameras have a baseline of $60~mm$ and a focal length of $2~mm$. The output consists of merged images from the two cameras, which are 640x240 resolution and have $60~fps$ frame rate.
	
	\begin{figure*}[t] 
		\centering
		\includegraphics[width=1\textwidth]{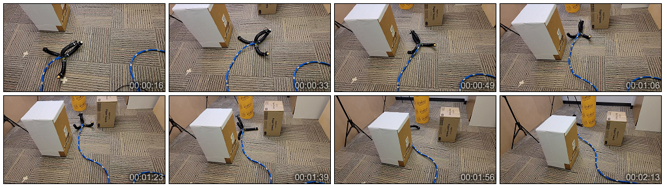}
		\caption{Teleoperation of the robot in an obstructed environment}
		\label{fig:teleoperation} 
	\end{figure*}
	
	\subsection{Generate Depth Information via Disparity Maps \label{subsec:DepthCalculation}}
	In teleoperation, sensory data from the robot in the field is crucial for ensuring the robot's utility and safety. Distance mapping is significant for operators as it provides spatial awareness, enabling them to avoid obstacles. Depth map estimation is a well-established area in robotics~\cite{biswas2022jomodevi}, and existing tools can be used to generate real-time maps using a stereo camera feed, as it is simpler to compute disparity and faster than other monocular methods~\cite{szeliski2022computer} (see Fig.~\ref{fig:depth}). The cameras were calibrated using a standard checkerboard method~\cite{opencv}, which generates distortion coefficients and camera matrix separately for each camera. These matrices are used to correct distortion before calculating disparity. We used the Block Matching approach for fast computation -- a block size of 21 was found to be the best. However, a disadvantage of this approach is that some regions' disparity is inaccurate due to plain texture and camera noise. Nevertheless, the effect of this is negligible as we are only interested in the approximate distance to objects in the field of view. Therefore, we found that coarse disparity calculation is sufficient, and large obstacles generate sufficient disparities for reasonably accurate estimations.
	
	\subsection{Topological Stability for Virtually Topple-Proof Robots\label{subsubsec:Toppleproof}}
	The robot design exhibits spatial symmetry, maintaining the same topology for any orientation when one limb is at a vertical position, as depicted in Fig.~\ref{fig:4_gaitpattern}. This feature enables the robot to navigate challenging terrains without reorienting, even when disoriented from its initial pose, as long as peripheral devices are symmetrically arranged. We demonstrate this feature by letting the robot fall from an elevated ground and execute the same gaits without reorientation, as shown in Fig.~\ref{fig:toppleproofexp}. The symmetric robot topology facilitates minimal kinematic transformations for re-mapping the limb configuration. Specifically, if the robot is toppled as illustrated in Fig.~\ref{fig:toppleproof}, all gaits are kinematically transformed into the respective limbs of the new robot orientation, except for $Limb_2$. However, the respective coordinate frames, $\{O_3\}$ and $\{O_4\}$, have not been mapped to the robot coordinate system, $\{O_R\}$ as depicted in Fig.~\ref{fig:toppleproof}B. To execute the transformed trajectories, all the gaits for $Limb_4$ must be defined on $\{O_4\}$ (where the \textit{XZ-plane} is perpendicular to the ground), and a $R_{z,3-4} \in \mathbb{SO}(3)$ is sufficient to transform $\{O_3\}$ to $\{O_4\}$. Thus, even after the topple, the system can identify $Limb_3$ as $Limb_4$ and executes the locomotion trajectories. Note that, $R_{z,3-4}$ are computed and predefined in the system for efficient use during re-mapping.
	
	\subsection{Orientation Correction of Limb Motion \label{subsubsec:TestingTurning}}
	Maintaining orientation is often critical in robot applications. Hence the ability to recover from a fall is an important feature. As shown in Fig.~\ref{fig:selfcorrection}, when the robot is disoriented, 
	$Limb_3$ rotates 
	in a counterclockwise direction (about $-Z$-axis).
	For example, if the robot is toppled to the left side, moving the joystick to the right corrects the orientation. Additionally, the robot can lift itself using a single limb to correct its orientation, enhancing its robustness.

	\section{Experimental Validation\label{sec:Experiments}}
	We experimentally validate the proposed teleoperation system through a series of experiments to evaluate the ability to navigate the robot in an unstructured environment using teleoperation with camera feedback. The complete set of experimental videos can be found at \url{https://youtu.be/8T402R8sZcg}

	\subsection{Experimental Setup\label{subsec:Experimental-Setup}}
	
	The soft 
	limbs
	use proportionally controlled air pressure for smooth bending operation. A large compressor provides a constant air pressure of $10~bar$ to digital proportional pressure regulators (ITV3050, SMC) connected to individual PMAs in each limb. The MATLAB, Simulink Desktop Real-Time model, and a NI PCI-6221 DAQ card generate proportional control voltage signals for the regulators. The same model also receives teleoperation commands from user inputs (analog voltages) through a NI PCI-6704 DAQ card.
	The terrain consists of a carpeted floor with varying friction coefficients due to the carpet pattern. To highlight the robot's capabilities, three obstacles were strategically placed for the teleoperation experiment. The topple-adaptation experiment was conducted on an elevated floor to simulate the falling of the robot, causing it to change orientation in a random direction. A person maneuvered the robot using the console, aided by a camera feed and depth map information.
	
	\subsection{Experiment Results and Discussion \label{subsubsec:DexterityController}}
	The robot is operated on a terrain with obstacles, as depicted in Fig.~\ref{fig:teleoperation}, where camera feedback is utilized to aid navigation. To improve the friction coefficient, the robot's hard plastic shell has friction tape added to its tip, allowing it to generate sufficient propulsion. Results confirm that the proposed 
	soft-limbed
	robot can support its own weight and carry sensing equipment while maintaining passive compliance. During navigation through a narrow gap smaller than the robot's size, limbs conform to the environment and passive compliance generates propulsion from the ground and static obstacles. Additionally, the deformation during actuation allows for undefined motions such as pushing away from obstacles (see Fig.~\ref{fig:toppleproofexp}. 
	The proportional control of gaits enhances navigation robustness, enabling proportional control of transverse speed, turning speed, and the angle at locations like narrow gaps, and improves the transition between gaits. The in-place turning capability of the robot is highlighted during navigation in narrow spaces where finely controlled rotational adjustments are required.
	
	The navigation of the robot is achieved by utilizing camera information
	and a disparity map, to assist the operator in identifying turning points and determining the heading direction (Fig.~\ref{fig:teleoperation}). While the camera may experience oscillations during locomotion, the stiffness of the top module, resulting from the structure and pneumatic pressure, quickly damps and stabilizes the oscillations. As a result, the teleoperation of the soft robot is demonstrated to be feasible and ready for autonomous navigation.
	
	In the second experiment (illustrated in Fig.~\ref{fig:toppleproofexp}), the topological stability feature of the robot is investigated. During the fall, the robot orientation changes randomly -- from the \textit{Blue-tip limb} to the \textit{Yellow-tip limb} (Fig.~\ref{fig:toppleproofexp}). To continue the teleoperation of the robot, the operator can identify the current orientation (i.e., the top limb) of the robot and 
	re-map
	the limb configuration from the system with the press of a button. Alternatively, 
	the operator can perform the self-orientation correction as described in Sec.~\ref{subsubsec:Toppleproof} without switching (refer to Fig.~\ref{fig:selfcorrection}).
	
	During the experiments, the robot exhibited stable movements even during contact with obstacles. Changing the CoG is an effective and reliable method to improve stability. Additionally, the bending of the robot's limbs facilitates the mounting of a forward-facing camera. 
	When turning, the robot tilts but remains stable. 
	We can compensate for this tilt by changing the bending direction of $Limb_1$, but the camera must then face the direction of movement.

	\section{Conclusions\label{sec:Conclusion}}
	We presented a teleoperation system designed for a soft mobile robot to navigate in an unstructured environment. The stable design locomotion gaits of the robot were introduced. Also, a console was developed to control the gaits proportionally and dynamically to navigate through obstructed environments using depth maps. The proposed 
	soft robot is spatially symmetric, and its limb functionality remapping feature ensures topological stability. This stability, which has not been demonstrated in the literature, is further enhanced by a self-orientation correction mechanism with a stable crawling gait. The experimental results demonstrated that the proportional control of dynamic gaits facilitates fine movements. Future work will focus on 
	deriving more efficient gaits 
	and implementing autonomous navigation with adaptive stability control for various terrain types and conditions.

	\bibliographystyle{IEEEtran}
	\bibliography{refs}

\end{document}